\documentclass[10pt,twocolumn,letterpaper]{article}

\usepackage{iccv}
\usepackage{times}
\usepackage{epsfig}
\usepackage{graphicx}
\usepackage{amsmath}
\usepackage{amssymb}

\usepackage{wrapfig}

\usepackage{multirow}

\usepackage[linesnumbered,ruled,vlined,noend]{algorithm2e}

\usepackage[accsupp]{axessibility} 

\newcommand{\boldm}[1] {\mathversion{bold}#1\mathversion{normal}}
\newcommand{\norm}[1]{\left\lVert#1\right\rVert}

\renewcommand\footnotemark{}

\usepackage[pagebackref=true,breaklinks=true,letterpaper=true,colorlinks,bookmarks=false]{hyperref}

\iccvfinalcopy 

\def\iccvPaperID{4311} 

\ificcvfinal\fi

\begin{document}

\title{Distribution-Aligned Diffusion for Human Mesh Recovery}

\author{Lin Geng Foo\textsuperscript{1}\thanks{~~\dag~Corresponding author}
~~~ Jia Gong\textsuperscript{1}
~~~ Hossein Rahmani\textsuperscript{2} ~~~ Jun Liu\textsuperscript{1\dag} \\
\textsuperscript{1}Singapore University of Technology and Design ~~ 
\textsuperscript{2}Lancaster University ~~ \\
{\tt\small \{lingeng\_foo,jia\_gong\}@mymail.sutd.edu.sg,  h.rahmani@lancaster.ac.uk, jun\_liu@sutd.edu.sg}\\ 
}

\maketitle
\ificcvfinal\thispagestyle{empty}\fi

\begin{abstract}
Recovering a 3D human mesh from a single RGB image is a challenging task due to depth ambiguity and self-occlusion, resulting in a high degree of uncertainty. Meanwhile, diffusion models have recently seen much success in generating high-quality outputs by progressively denoising noisy inputs. Inspired by their capability, we explore a diffusion-based approach for human mesh recovery, and propose a Human Mesh Diffusion (HMDiff) framework which frames mesh recovery as a reverse diffusion process. We also propose a Distribution Alignment Technique (DAT) that infuses prior distribution information into the mesh distribution diffusion process, and provides useful prior knowledge to facilitate the mesh recovery task. Our method achieves state-of-the-art performance on three widely used datasets. Project page: https://gongjia0208.github.io/HMDiff/.
\end{abstract}

\section{Introduction}

Monocular 3D human mesh recovery (HMR), where the 3D mesh vertex locations of a human are predicted from a single RGB image, is an important task with applications across virtual reality \cite{huang2017towards}, sports motion analysis \cite{aggarwal1999human}, and
healthcare \cite{treleaven20073d}.
The field has received a lot of attention in recent years \cite{cho2022cross,choi2020pose2mesh,lin2021end,lin2021mesh,choi2022learning,khirodkar2022occluded}, which has led to significant progress, but monocular 3D HMR still remains very challenging. 
Human body shapes are not only complex and contain many fine details, but also inherently exhibit \textit{depth ambiguity} (when recovering 3D information from single 2D images) and \textit{self-occlusion} (where body parts can be occluded by other body parts) \cite{kocabas2021pare,cho2022cross,tian2022recovering}.
In particular, the depth ambiguity and self-occlusion in this task often bring much uncertainty to the recovery of 3D mesh vertices,
and places a huge burden on the model to handle this inherent uncertainty 
\cite{lee2021uncertainty,moon2020i2l,choi2022learning,khirodkar2022occluded}.

At the same time, denoising diffusion probabilistic models (\textit{diffusion models}) \cite{ho2020denoising, sohl2015deep} have recently seen much success in generative tasks, 
such as image \cite{lugmayr2022repaint}, video \cite{singer2022make} and text \cite{li2022diffusion} generation, where they have been capable of producing highly realistic and good-quality samples.
Specifically, diffusion models \cite{ho2020denoising, sohl2015deep} progressively ``denoise" a noisy input -- which is uncertain -- into a high-quality output from the desired data distribution (e.g., natural images),
through estimating the gradients of the data distribution \cite{song2019generative} (also known as the score function).
This progressive denoising helps break down the large gap between distributions (i.e., from a highly uncertain and noisy distribution to a desired target distribution), into smaller intermediate steps \cite{song2019generative}, which assists the model in converging towards generating the target data distribution smoothly.
This gives diffusion models a strong ability to
recover high-quality outputs from uncertain and noisy input data.

For the monocular 3D mesh recovery task, we also seek to recover a high-quality mesh prediction from uncertain and noisy input data, 
and so we leverage diffusion models to effectively tackle this task.
To this end, we propose a novel diffusion-based framework for monocular 3D human mesh recovery, called Human Mesh Diffusion (\textbf{HMDiff}), where we frame the mesh recovery task as a \textit{reverse diffusion process} which recovers a high-quality mesh by progressively denoising noisy and uncertain input data.

Intuitively, in our HMDiff approach, we can regard the mesh vertices as particles in the context of thermodynamics.
At the start, the particles (representing the ground truth mesh vertices) are systematically arranged to form a high-quality human mesh. Then, these particles gradually disperse throughout the space and degrade into noise, leading to high uncertainty.
This process (i.e., particles becoming more dispersed and noisy) is the \textit{forward diffusion process}.
Conversely, for human mesh recovery, we aim to perform the opposite of this process, i.e., the \textit{reverse diffusion process}.
Starting from a noisy and uncertain input distribution $H_K$, we aim to progressively denoise and reduce the uncertainty of the input to obtain a target human mesh distribution containing high-quality samples.

Correspondingly, our HMDiff framework consists of both the forward process and the reverse process.
Specifically, the forward process is performed during training to generate samples of intermediate distributions that are used as step-by-step supervisory signals to train our diffusion model $g$.
On the other hand, the reverse process is a crucial part of our mesh recovery pipeline, which is used during both training and testing.
In the reverse process, we first initialize a noisy distribution $H_K$, and use our diffusion model $g$ to progressively transform $H_K$ into a high-quality human mesh distribution ($H_0$) over $K$ diffusion steps.

However, it can be difficult to recover a high-quality mesh distribution from the noisy input distribution $H_K$, by using only the standard diffusion process \cite{ho2020denoising}.
This is due to the high complexity of the dense 3D mesh structure \cite{choi2020pose2mesh,zheng2022lightweight,lin2021end}, which makes it difficult to directly produce accurate 3D mesh outputs with a single RGB image as input.
Thus, as shown in many previous works \cite{choi2020pose2mesh,yu2021skeleton2mesh,pavlakos2018learning,lassner2017unite,choi2022learning,khirodkar2022occluded,omran2018neural,zanfir2021neural} on HMR, it is essential to also leverage the prior knowledge extracted via pre-trained extractors (e.g., pose information, segmentation maps) to guide the 3D mesh recovery process.

Therefore, we further propose a Distribution Alignment Technique (\textbf{DAT}), where we infuse prior distribution information into the reverse mesh distribution diffusion process, effectively leveraging prior knowledge to facilitate the 3D mesh recovery task.
The prior distribution (e.g., a pose heatmap encoding rich semantic and uncertainty information \cite{luo2021rethinking,kundu2022uncertainty,han2022single}) is extracted from the input image with a pose estimator \cite{wang2020deep}, and is used to guide the initial diffusion steps towards the diffusion target $H_0$ via a modified diffusion process.
However, there exists challenges in infusing the prior distribution information into the diffusion steps.
For instance, the mesh distribution $H_k$ at each $k$-th step of the process has its own characteristics, such that the $k$-th step of the reverse diffusion process is trained specifically (conditioned on step index $k$) to bring samples from the distribution $H_k$ to the distribution $H_{k-1}$.
Thus, if we directly modify the sample $h_k$ at step $k$ (i.e., $h_k \sim H_k$) without careful consideration, it can potentially cause the sample $h_k$ to deviate from the distribution $H_k$, which disrupts the diffusion process and leads to sub-optimal performance.

Hence, in order to infuse prior distribution information to the mesh distribution diffusion process without disrupting it, our DAT does not directly modify $h_k$.
Instead, DAT aligns the diffusion steps towards the  prior distribution by taking the $k$ remaining diffusion steps into account, carefully updating $h_k$ such that the eventual prediction after $k$ diffusion steps is pulled closer to the diffusion target.
Overall, DAT infuses the diffusion process with prior distribution information without disrupting the diffusion process, leading to faster convergence (i.e., fewer diffusion steps) and better performance.

\section{Related Work}

\textbf{Human Mesh Recovery (HMR)} aims to recover the 3D human mesh from a given input.
Traditionally, HMR has been successful with the aid of information from depth sensors \cite{shin20193d,newcombe2011kinectfusion} and inertial measurement units \cite{huang2018deep,von2018recovering}.
Recently, there has been much research attention on the monocular setting \cite{tian2022recovering}, which is more convenient and widely applicable.
Yet, monocular HMR is very challenging due to depth ambiguity, occlusions and complex pose variations \cite{lin2021end,tian2022recovering,biggs20203d}.
Many works \cite{kanazawa2018end,pavlakos2018learning,omran2018neural,kocabas2021pare,guan2009estimating,sigal2007combined,biggs20203d} tackle monocular HMR via a \textit{model-based} approach, where networks are trained to regress the parameters of a human parametric model (e.g., SMPL \cite{loper2015smpl}, SCAPE \cite{anguelov2005scape}).
To further improve performance, some works propose to leverage various forms of prior knowledge as guidance, 
including pose \cite{choi2020pose2mesh,yu2021skeleton2mesh},
pose heatmaps \cite{choi2022learning,khirodkar2022occluded,lassner2017unite},
or segmentation maps \cite{omran2018neural,zanfir2021neural}.
Recently, some works adopt a \textit{model-free} approach \cite{choi2020pose2mesh,lin2021end,lin2021mesh,kolotouros2019convolutional,moon2020i2l,cho2022cross}, where the full 3D human body shape is directly predicted.
In this line of work, several models have been explored, including Convolutional Neural Networks (CNNs) \cite{moon2020i2l}, Graph Convolutional Networks (GCNs) \cite{kolotouros2019convolutional,choi2020pose2mesh} and Transformers \cite{lin2021end,lin2021mesh,cho2022cross}.
Differently, here we explore a diffusion-based framework to handle the uncertainty in HMR.
To the best of our knowledge, this is the first work to use diffusion models to tackle monocular HMR.
Our HMDiff framework effectively recovers human 3D mesh and achieves state-of-the-art performance.

\textbf{Denoising Diffusion Probabilistic Models (Diffusion Models)} \cite{sohl2015deep,ho2020denoising} effectively enable us to learn to sample from a desired data distribution, 
by iteratively ``denoising'' random noise into a high-quality sample from the desired data distribution
through estimating the gradients (i.e., score function) of the data distribution \cite{song2019generative,chung2023diffusion}.
Diffusion models have been effective at image generation \cite{ho2020denoising,song2021denoising,foo2023aigc},
and have been explored for various other generation tasks such as 
video generation \cite{singer2022make}, 
and text generation \cite{li2022diffusion}. 
Recently, several works also explore applying diffusion models in prediction tasks \cite{gong2023diffpose,gu2022stochastic,peng2023diffusion} and image-based inverse problems \cite{chung2023diffusion,kawar2022denoising,song2023pseudoinverseguided}.
In contrast to these studies, monocular HMR presents a more difficult challenge, requiring a dense (mesh) output with only a single input image.
Thus, to simplify the task of monocular HMR, we take inspiration from previous works \cite{chung2023diffusion,kawar2022denoising,song2023pseudoinverseguided} that estimate the posterior, and adopt a similar approach to guide the initial stages of the diffusion process with our DAT. 
Our DAT aligns the initial mesh distribution towards an extracted prior pose distribution (bridged by a mesh-to-pose function), resulting in faster convergence and better performance.

\section{Background on Diffusion Models}
\label{sec:background}

Overall, diffusion models \cite{ho2020denoising,song2021denoising} 
are probabilistic generative models that learn to transform random noise $h_{K} \sim \mathcal{N}(\mathbf{0},\mathbf{I})$ into a desired sample $h_0$ by denoising $h_{K}$ in a recurrent manner.
Diffusion models have two opposite processes: the \textit{forward process} and the \textit{reverse process}.

Specifically, in the forward process, a ``ground truth" sample $h_0$ with low uncertainty is gradually diffused over $K$ steps $(h_0 \rightarrow h_{1} \rightarrow ... \rightarrow h_{K})$ towards becoming a sample $h_K$ with high uncertainty. 
Samples are obtained from the intermediate steps along the way, which are used during training as the step-by-step supervisory signals for the diffusion model $g$.
To start the reverse process, noisy and uncertain samples $h_K$ are first initialized according to a standard Gaussian distribution.
Then, the diffusion model $g$ is used in the reverse process ($h_K \rightarrow h_{K-1} \rightarrow ... \rightarrow h_{0}$) to progressively reduce the uncertainty of $h_K$ and transform $h_K$ into a sample with low uncertainty ($h_0$).
The diffusion model $g$ is optimized using the samples from intermediate steps (generated in the forward process), which guide it to smoothly transform the noisy and uncertain samples $h_K$ into high-quality samples $h_0$.
We go into more detail below.

\textbf{Forward Process.}
Firstly, the forward diffusion process generates a set of intermediate noisy samples $\{h_k\}^{K-1}_{k=1}$ that will be used to aid the diffusion model in learning the reverse diffusion process during training.
Since Gaussian noise is added between each step, we can formulate the posterior distribution $q(h_{1:K}|h_0)$ as:

\begin{footnotesize}
\setlength{\abovedisplayskip}{0pt}%
\label{eq: original diffusion}
\begin{align}
q(h_{1:K}|h_{0}) &:=  \prod_{k=1}^{K} q(h_{k}|h_{k-1})  \\
\label{eq:diffusion_hk_forward_step}
q(h_{k}|h_{k-1}) &:= \mathcal{N}_{pdf} \big(h_k \big| \sqrt{\frac{\alpha _k}{\alpha _{k-1}}}h_{k-1},(1-\frac{\alpha _k}{\alpha _{k-1}})\mathbf{I} \big),  
\end{align}
\end{footnotesize}
where $\mathcal{N}_{pdf}(h_k| \cdot)$ is the likelihood of sampling $h_k$ conditioned on the given parameters, while $\alpha _{1:K} \in (0,1]^K$ is a fixed decreasing sequence that controls the noise scaling at each diffusion step.
We can then formulate the posterior $q(h_k|h_0)$ for the diffusion process from $h_0$ to step $k$ as:

\begin{footnotesize}
\setlength{\abovedisplayskip}{0pt}
\begin{align}
q(h_k|h_0) :=& \int q(h_{1:k}|h_0) \text{d} h_{1:k-1} \notag \\
\label{eq:diffusion_hk_from_h0}
=& \mathcal{N}_{pdf}(h_k | \sqrt{\alpha _k}h_0, (1-\alpha _k)\mathbf{I}).
\end{align}
\end{footnotesize}

Hence, we can express $h_k$ as a linear combination of  the source sample $h_0$ and random noise $z$, where each element of $z$ is sampled from $\mathcal{N}(\mathbf{0},\mathbf{I})$, as follows:

\begin{footnotesize}
\setlength{\abovedisplayskip}{4pt}
\begin{align}
\label{eq: original diffusion linear function} 
h_k =\sqrt{\alpha _k}h_0 + \sqrt{(1-\alpha _k)} z.
\end{align}
\end{footnotesize}

Thus, by setting a decreasing sequence $\alpha _{1:K}$ such that $\alpha_K \approx 0$, the distribution of $h_K$ will converge to a standard Gaussian ($h_K \sim \mathcal{N}(\mathbf{0},\mathbf{I})$).
Intuitively, this implies that the source signal $h_0$ will eventually be corrupted into Gaussian noise $h_K$,
which matches with the non-equilibrium thermodynamics phenomenon of the diffusion process \cite{sohl2015deep}.
This facilitates the training of the reverse process, as the generated samples $\{h_k\}_{k=1}^K$ effectively bridge the gap between the standard Gaussian noise $h_K$ and the source sample $h_0$.

\textbf{Reverse Process.}
Next, to approximate the reverse diffusion process which transforms Gaussian noise $h_K$ to a high-quality sample $h_0$, a diffusion model $g$ (which is often a deep network) is optimized using the generated samples $\{h_k\}_{k=1}^K$ and the source sample $h_0$.
The diffusion model $g$ can be interpreted to be a score-based model \cite{song2019generative} that estimates the \textit{score function} $\nabla_{h} \log p(h)$ of the data distribution, where the iterative steps are performing denoising score matching \cite{vincent2011connection} over multiple noise levels.
The standard formulation of the reverse diffusion step by DDPM \cite{ho2020denoising} can be formally expressed as:

\begin{footnotesize}
\setlength{\abovedisplayskip}{0pt}
\begin{align}
\label{eq:reverse diffusion} 
h_{k-1} = \sqrt{ \frac{\alpha_k}{\alpha_{k-1}}} \big(h_k - \frac{\alpha_{k-1} - \alpha_k}{\alpha_{k-1} \sqrt{1 - \bar{\alpha}_k}}g(h_k,k) \big) + \sigma_k z_k,
\end{align}
\end{footnotesize}
where $\sigma_k$ is a hyperparameter and $z_k$ is Gaussian noise.

Therefore, during inference, Gaussian noise $h_K$ can be sampled, and the reverse diffusion step introduced in Eq.~\ref{eq:reverse diffusion} can be recurrently performed, allowing us to generate a high-quality sample $h_0$ with the trained diffusion model $g$.

\section{Method}
\label{sec:Method}

\begin{figure*}[th]
  \centering
  \includegraphics[width=1\linewidth]{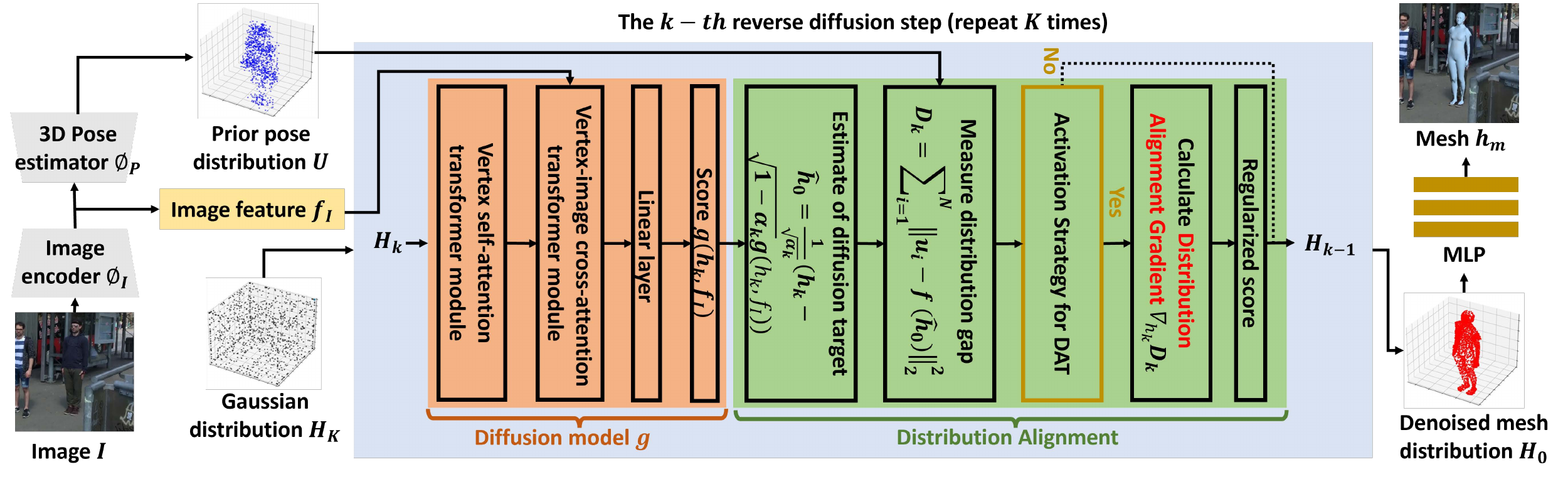}
  \caption{
    Illustration of the proposed Human Mesh Diffusion (HMDiff) framework with the Distribution Alignment Technique (DAT).
    Given an RGB image $I$, we first extract an image feature $f_I$ and a prior distribution $U$, by using pre-trained models $\phi_I$ and $\phi_P$. 
    Then, a Transformer-based diffusion model $g$ takes in $U$, $f_I$ and noise from a Gaussian distribution $H_{K}$, and iteratively performs $K$ denoising diffusion steps to eventually obtain a denoised distribution $H_0$.    
    We use our DAT technique to guide the diffusion process by computing a Distribution Alignment Gradient, which effectively infuses prior distribution information to guide the denoising steps.   
    Lastly, after obtaining the high-quality mesh distribution $H_0$, we take the center of $H_0$ and feed it into an MLP to obtain the final prediction $h_m$. 
    }
  \label{fig:overall_pipeline}
\end{figure*}

In this section, we first formulate our HMDiff framework (in Sec.~\ref{sec:reconstruction_pipeline}). An overview of our framework is depicted in Fig.~\ref{fig:overall_pipeline}.
Then, in Sec.~\ref{sec: Distribution Alignment Technique}, we propose DAT to infuse prior distribution information into the diffusion process.
Lastly, we design a diffusion network for HMR in Sec.~\ref{sec: architecture} that can effectively model the relationship between vertices.

\subsection{Human Mesh Diffusion (HMDiff)}
\label{sec:reconstruction_pipeline}

Monocular 3D HMR is a very challenging task, due to the inherent depth ambiguity in recovering 3D information from single 2D images, as well as self-occlusion where some body parts may be occluded by other body parts. These issues often result in high uncertainty during 3D mesh recovery \cite{kocabas2021pare,cho2022cross,tian2022recovering,lee2021uncertainty}.
Thus, in order to alleviate the uncertainty, we propose to leverage diffusion models, which have a strong capability in recovering high-quality outputs from noisy and uncertain data.

To this end, we propose HMDiff, a diffusion-based framework for HMR, which consists of a \textit{forward process} and a \textit{reverse process}.
The forward process gradually adds noise and uncertainty to the ground truth mesh samples, eventually corrupting the mesh vertices into Gaussian noise. 
On the other hand, the reverse process learns to \textit{reverse the effects of the forward process}, and learns to model the step-by-step reduction of noise and uncertainty, gaining the ability to progressively denoise the noisy inputs to recover a high-quality mesh.
Specifically, we frame the HMR task as a reverse diffusion process, while the forward process plays a crucial role during training, as described below.

\textbf{Forward Process.}
To train the diffusion model $g$ to bridge the large gap between the target human mesh distribution $H_0$ and the standard Gaussian distribution $H_K$, we first generate a set of intermediate distributions $\{H_1, H_2,..., H_{K-1}\}$ in the forward process.
We obtain the intermediate distributions via Eq.~\ref{eq: original diffusion linear function}, and use them as step-by-step supervisory signals to train the diffusion model $g$.
Throughout the forward process, we apply noise to the vertex coordinates while keeping the topology between vertices fixed.
To be more precise, we set the topology according to a predefined adjacency matrix (following \cite{cho2022cross,choi2020pose2mesh}) and keep it fixed.

\textbf{Reverse Process.}
In the reverse process, we aim to recover an accurate and high-quality human mesh {distribution} $H_0$ from a noisy and uncertain input $H_K$ (initialized as a standard Gaussian distribution) through a step-by-step process where the uncertainty is reduced.
To achieve this, we design a diffusion model $g$, which predicts the score function in each $k$-th reverse diffusion step, conditioned on step index $k$.
Then, we can use diffusion model $g$ to perform reverse diffusion by recurrently taking reverse diffusion steps.
{Specifically, the reverse process consists of $K$ steps that produce mesh distributions $p_k (h_k)$ at each $k$-th step, which are trained such that $p_k (h_k)$ matches $q(h_k | \cdot)$ of the forward process (i.e., Eq.~\ref{eq:diffusion_hk_forward_step} and Eq.~\ref{eq:diffusion_hk_from_h0}) at the $k$-th step.}
Overall, the reverse process is trained to produce $p_0(h_0)$ at step 0, which corresponds to the high quality mesh distribution $H_0$.

However, rather than taking the standard steps (Eq.~\ref{eq:reverse diffusion}) for reverse diffusion, here we build our mesh reverse diffusion process based on another formulation \cite{song2021denoising}.
This is due to two reasons.
Firstly, this formulation greatly improves the efficiency of the diffusion process by enabling us to jump across multiple diffusion steps at once, which reduces the number of diffusion steps required during inference while also maintaining the performance of the standard diffusion process \cite{ho2020denoising}.
Secondly, this formulation consists of a way to jump over $k$ steps at once (in order to facilitate the skipping of steps to improve efficiency \cite{song2021denoising}) -- this offers a convenient way to jump over $k$ steps to obtain an early prediction and facilitate our DAT, which will be explained in detail in Sec.~\ref{sec: Distribution Alignment Technique}.
Specifically, we adopt the following formulation for the $k$-th reverse diffusion step:

\begin{footnotesize}
\setlength{\abovedisplayskip}{0pt}%
\begin{align}
\label{eq:DDIM reverse diffusion} 
h_{k-1} =  \sqrt{1 - \alpha_{k-1} - \sigma_k^2} \cdot g(h_k,k) + \sqrt{\alpha_{k-1}} \cdot \hat{h}_0 + \sigma_k z_k, 
\end{align}
\end{footnotesize} 
where $h_k$ is a sample at the $k$-th step ($h_k \sim H_k$), $\alpha_k,\sigma_k$ are hyperparameters and $z_k$ is standard Gaussian noise.
We remark that this formulation for reverse diffusion is widely applicable, and as noted in \cite{song2021denoising}, is equivalent to DDPM \cite{ho2020denoising} when we set $\sigma_k = \sqrt{(1-\alpha_{k-1})/(1-\alpha_k)} \sqrt{1 - \alpha_k/\alpha_{k-1}}$ for all $k$.
Following \cite{song2021denoising}, we can further define $\hat{h}_0$ as:

\begin{footnotesize}
\setlength{\abovedisplayskip}{0pt}%
\setlength{\belowdisplayskip}{0pt}%
\begin{align}
    \label{eq: estimate hat h_0}
    \hat{h}_0 = \frac{1}{\sqrt{\alpha_k}} (h_k - \sqrt{1 - \alpha_k} \cdot g (h_k,k) ).   
\end{align}
\end{footnotesize}

Intuitively, $\hat{h}_0$ represents an estimation of the target ($h_0$) by using the sample at the current step ($h_k$) and ``reverse diffusing'' in one large step, without taking $k$ steps. 
Eq.~\ref{eq: estimate hat h_0} helps to facilitate the jump over multiple steps using the DDIM acceleration technique \cite{song2021denoising} during testing, which greatly improves efficiency.
Furthermore, to facilitate the reverse diffusion process, we also extract an image feature $f_I$ from the image, and feed it to the diffusion model $g$ at each step; more details can be found in Sec.~\ref{sec: architecture}.

Overall, after recurrently applying Eq.~\ref{eq:DDIM reverse diffusion} on $h_K$ 
for $K$ iterations, we can get $h_0$, where $h_0 \sim H_0$ and
$H_0$ is the denoised (high-quality) mesh distribution.
In practice, we can run this reverse diffusion process in parallel $N$ times, and obtain $N$ samples of $h_0$ to represent $H_0$.

However, even with the reverse diffusion process introduced above, it can still be difficult to predict a highly accurate 3D mesh, since
3D mesh distributions are highly dense and complex \cite{choi2020pose2mesh,yu2021skeleton2mesh,pavlakos2018learning,lassner2017unite,choi2022learning,khirodkar2022occluded,omran2018neural,zanfir2021neural}, making it challenging to produce them with a single RGB image as input.
Thus, we propose a way to extract prior distribution information from the image (from pre-trained extractors) and use it to guide the mesh distribution diffusion process, as described next.

\subsection{Distribution Alignment Technique (DAT)}
\label{sec: Distribution Alignment Technique}

We develop DAT to infuse prior distribution information to the reverse mesh distribution diffusion process.
Specifically, in order to bridge the large gap between the Gaussian distribution ($H_K$) and the desired diffusion target $H_0$, we extract and use a prior distribution $U$ that strongly correlates to $H_0$, 
e.g., in this paper, $U$ is a pose heatmap which contains rich semantic and uncertainty information \cite{luo2021rethinking,kundu2022uncertainty,han2022single} of the input image.
Since $U$ strongly correlates with $H_0$, we seek to align our initial distribution $H_K$ towards the prior distribution $U$ in the initial diffusion steps.
Intuitively, this \textit{narrows down the target} of the mesh distribution diffusion process to the neighbouring area around $U$,
where the rest of the diffusion process can further reduce the uncertainty to eventually obtain the desired target mesh distribution $H_0$. 
This greatly reduces the difficulty of the 3D mesh recovery process.

However, we face a challenge in infusing the distribution information $U$ into the mesh distribution diffusion steps, because it can potentially disrupt the diffusion process.
More precisely, the distribution at each $k$-th step of the diffusion process has its own characteristics, e.g., the diffusion model is conditioned on the diffusion step index $k$ to specifically bring the sample $h_k$ (from distribution $H_k$) towards $h_{k-1}$ (from distribution $H_{k-1}$).
Therefore, if we infuse information by directly modifying the sample $h_k$ at each step, it might cause the sample $h_k$ to deviate from distribution $H_k$, thereby derailing the diffusion process and disrupting it.

Hence, in order to infuse prior distribution information to the mesh distribution diffusion process without disrupting it, we \textit{do not directly modify $h_k$}, but instead indirectly align it to $U$ via a Distribution Alignment Gradient.
Specifically, to compute the Distribution Alignment Gradient for the $k$-th diffusion step, we first estimate an early prediction of the diffusion target ($\hat{h}_0$) while conditioned on $k$, i.e., taking into account the $k$ remaining diffusion steps.
Next, we compare the early estimate $\hat{h}_0$ with the prior distribution information, and compute the Distribution Alignment Gradient by minimizing their gap.
As a result, the computed gradients give guidance on how to align $h_k$, \textit{such that after $k$ diffusion steps the prediction $\hat{h}_0$ is pulled closer to $U$ (and the target $H_0$)}, which does not disrupt the diffusion process.
We introduce our DAT in detail below, and an illustration can be seen in Fig.~\ref{fig:overall_pipeline}.

\textbf{Theoretical Formulation.}
First, we formally introduce some definitions.
We are given a prior distribution $U$ and Gaussian noise $h_{K} \sim H_{K}$, and want to produce an output $h_0 \sim H_0$ after $K$ iterations of reverse diffusion.
We also define the relationship between the samples from the target mesh distribution $H_0$ and the prior distribution $U$ as: $f(h_0) + n = u$, where $f$ is a function (e.g., a linear mesh-to-pose function if $U$ is a human pose distribution), $u \sim U$, and $n$ is some error which is defined since the prior distribution $U$ can be imprecise and uncertain.
Moreover, since the diffusion model $g$ can be interpreted to be a learned score estimator \cite{song2019generative,ho2020denoising} to estimate the score function (distribution gradient) $\nabla_{h_k} \log p_k (h_k) $,
the reverse diffusion step in Eq.~\ref{eq:DDIM reverse diffusion} can be reformulated as:

\begin{footnotesize}
\setlength{\abovedisplayskip}{0pt}
\setlength{\belowdisplayskip}{4pt}
\begin{align}
\label{eq:reverse diffusion score function}
    h_{k-1} =& \sqrt{1- \alpha_{k-1} - \sigma_k^2}  \cdot \nabla_{h_k} \log p_k (h_k) + \sqrt{\alpha_{k-1}} \cdot \hat{h}_0 + \sigma_k z_k 
\end{align}
\end{footnotesize}

Intuitively, this shows that the evolution between each step (e.g., $h_k \rightarrow h_{k-1}$) is dependent on the gradient of the data distribution $\nabla_{h_k} \log p_k (h_k )$ -- which is called the score function  -- that pushes $h_k$ to $h_{k-1}$.
Importantly, here we want this gradient to take into account the prior distribution $U$ and align towards it. 
In other words, instead of estimating the gradient of the data distribution $\nabla_{h_k} \log p_k (h_k )$, we want an estimate that is conditioned on $U$, i.e., $\nabla_{h_k} \log p_k (h_k | U)$, which allows us to \textit{effectively infuse prior distribution information into the mesh distribution diffusion process}.

Next, we aim to find a way to compute $\nabla_{h_k} \log p_k (h_k | U)$.
From Bayes' rule, we can get
$\nabla_{h_k} \log p(h_k| U) = \nabla_{h_k} \log p_k (h_k) + \nabla_{h_k} \log p(U| h_k)$, where we already have the first term (i.e., the original score function).
Hence, in order to compute the gradient conditioned on $U$, we only need to find a way to compute the second term, i.e., $\nabla_{h_k} \log  p(U | h_k)$.
To achieve this, intuitively we need to find a differentiable function that connects $U$ and $h_k$ and compute its gradient, i.e., minimizing the gap between the noisy $h_k$ and the prior distribution $U$.
Such a gradient will directly update $h_k$ to be closer to $U$, thereby infusing prior information directly into $h_k$ at each step $k$.

\textbf{Why we use $\hat{h}_0$.}
However, it is not feasible to apply the above-mentioned gradient to \textit{directly modify} $h_k$.
This is because the mesh distribution $H_k$ at every diffusion step $k$ has its own characteristics; hence, to learn how to bring a sample specifically from $H_k$ to $H_{k-1}$, the diffusion model $g$ is trained conditioned on the step index $k$ (see Eq.~\ref{eq:reverse diffusion}).
Therefore, if we forcefully align $h_k$ to be closer to $U$, it can deviate away from $H_k$ and be different from what the diffusion model was trained to denoise at step $k$, and thus disrupt the diffusion process.
Hence, we also do not want to directly modify $h_k$.

Next, we observe that our aim is to eventually predict an accurate mesh sample $h_0$ -- \textit{not to align $h_k$ itself to be close to $U$}.
Thus, we propose an indirect approach to align $h_k$, which does not involve directly modifying $h_k$ to be more similar to $U$. 
Specifically, for every $h_k$, we can first compute an early estimate of $h_0$ (i.e., $\hat{h}_0$) as an intermediate prediction via Eq.~\ref{eq: estimate hat h_0} which is conditioned on $k$. 
Next, we can compute some gradients that will make $\hat{h}_0$ into a better prediction (i.e., by minimizing the gap with $U$), then propagate those gradients back from $\hat{h}_0$ to $h_k$.
As a result, these gradients give guidance on how to align $h_k$, \textit{such that after $k$ diffusion steps the prediction $\hat{h}_0$ is pulled closer to $U$ (and the target $H_0$)}, which fulfills our objective without disrupting the diffusion process.

Specifically, we first estimate $\hat{h}_0$ as a function of $h_k$ (i.e., in Eq.~\ref{eq: estimate hat h_0}); we slightly abuse notation to denote this as $\hat{h}_0 (h_k)$, to make it clear that $\hat{h}_0$ is estimated as a function of $h_k$.
Using $\hat{h}_0 (h_k)$, we can approximate $p(U | h_k)$ by computing $p(U | \hat{h}_0 (h_k) )$, which can be used to compute the gradients to update $h_k$.

\textbf{Distribution Alignment Gradient.} 
Next, we show how the gradient $\nabla_{h_k} \log p(U | \hat{h}_0 (h_k))$ can be computed in practice to obtain a Distribution Alignment Gradient that can infuse prior distribution information into the mesh distribution diffusion process.  
Specifically, we can interpret $-\log p(U | \hat{h}_0)$ to be the negative log-posterior of observing $U$ given $\hat{h}_0$, which tends to have higher magnitude as the difference between $U$ and $\hat{h}_0$ gets larger, and we can derive a gradient by trying to minimize the gap $D_k$ between them.
However, $U$ is a pose distribution, while $\hat{h}_0$ is in mesh format, making it difficult to directly compute the gap $D_k$.
Thus, we introduce a mesh-to-pose function $f$ to map $\hat{h}_0$ to a corresponding pose, and minimize the gap between $U$ and $f(\hat{h}_0)$.
To efficiently calculate $D_k$, we sample $N$ elements from $U$ (i.e., $u \sim U$), and calculate the sum of $L_2$ norms between $u$ and $f (\hat{h}_0)$, as follows:

\begin{footnotesize}
\setlength{\abovedisplayskip}{0pt}%
\begin{align}
    \nabla_{h_k} \log p(U| h_k) \approx  -  \nabla_{h_k} D_k,
\end{align}
\setlength{\abovedisplayskip}{0pt}%
\setlength{\belowdisplayskip}{0pt}%
\begin{align}
    \label{eq:Dk}
    D_k = \sum_{i=1}^{N} || u_i - f(\hat{h}_0 (h_k)) ||^2_2,\,\, u_i \sim U,
\end{align}
\end{footnotesize}
where $- \nabla_{h_k} D_k$ is our Distribution Alignment Gradient.
In practice, we also add a hyperparameter $\gamma$ as a coefficient to $-\nabla_{h_k} D_k$.
This Distribution Alignment Gradient aligns the diffusion steps towards the prior distribution $U$ (and the diffusion target $H_0$) at the initial stages of the diffusion process, greatly reducing the difficulty of the 3D mesh recovery process.
However, the role of DAT diminishes in importance towards the end of the diffusion process, after having aligned $h_k$ towards $U$ and $H_0$. 
Thus, we design an Activation Strategy to decide when we apply the gradient, as explained next.

\textbf{Activation Strategy for DAT.}
Our DAT aims to infuse prior distribution information into the mesh distribution diffusion process, and align the initial Gaussian distribution $H_K$ towards a  prior distribution $U$, which is close to the target $H_0$.
However, as the extracted prior distribution $U$ still contains uncertainty and noise, we do not want to overly align $H_k$ towards $U$ in the later parts of the diffusion process, when $H_k$ is relatively more accurate and certain.
Hence, we propose activating (or deactivating) DAT based on the measured gap ($D_k$) between $h_k$ and $U$, where we deactivate it when $H_k$ converges to a more compact and high-quality distribution.

Specifically, at the start, we measure the initial gap $D_K$ (between $U$ and $h_K$) based on Eq.~\ref{eq:Dk}.
Then, at each step $k$, we measure the gap $D_k$, and calculate the relative distribution gap value $R_k = \frac{D_k}{D_{K}}$.
At the start (when $k=K$), $R_k = 1$, and $R_k$ starts to shrink as $H_k$ gets aligned to $U$.
We activate DAT as long as the relative distribution gap value $R_k$ is more than a specified threshold $r$.
On the other hand, when $R_k < r$ for some $k$-th step, we terminate DAT for the steps thereafter, since the distribution $H_k$ has already become rather aligned to $U$ and the target $H_0$.

In summary, the detailed reverse diffusion process with our DAT can be seen in Algorithm \ref{alg:dps_gauss}.

\begin{algorithm}[t]
\caption{The DAT Reverse Diffusion Process}
\label{alg:dps_gauss}
\SetAlgoLined
\SetNoFillComment
\SetKwInput{KwData}{Input}
\scriptsize
\KwData{
prior distribution $U$, number of samples $N$, decreasing sequence $\alpha_{1:K}$, sequence $\sigma_{1:K}$, diffusion model $g$, threshold $r$, DAT weight $\gamma$. }
Sample a noise $h_K$, where $h_K \sim \mathcal{N}(\mathbf{0},\mathbf{I})$

{$act = True$}\tcp{Initialize activation status}
\For(\tcp*[h]{Reverse diffusion process}){$k= K$ to $1$}
{$s \gets g (h_k, k)$ \tcp{Estimate score}
$\hat{h}_0 \gets \frac{1}{\sqrt{\alpha_k}} (h_k - \sqrt{1 - \alpha_k} \cdot s )$ \tcp{Estimate $\hat{h}_0$}
$D_k \gets \sum_{i=1}^{N}  || u_i - f(\hat{h}_0) ||^2_2$\tcp{Measure the gap}
$R_k \gets \frac{D_{k}}{D_{K}}$\tcp{Measure the relative gap}
\uIf(\tcp*[h]{with DAT}){$R_k \geq r$ and {$act == True$}}
{$h_{k-1} \gets \sqrt{1 - \alpha_{k-1} - \sigma_k^2} \cdot s + \sqrt{\alpha_{k-1}} \cdot \hat{h}_0 - \gamma \nabla_{h_k} D_k + \sigma_k z_k$}
\Else(\tcp*[h]{without DAT})
{
{$act \gets False$}
$h_{k-1} \gets \sqrt{1 - \alpha_{k-1} - \sigma_k^2} \cdot s + \sqrt{\alpha_{k-1}} \cdot \hat{h}_0 + \sigma_k z_k$}
} 
\end{algorithm}

\subsection{Network Architecture}
\label{sec: architecture}

As shown in Fig.~\ref{fig:overall_pipeline}, our full pipeline consists of a CNN backbone $\phi_I$ with a 3D pose estimator head $\phi_P$, a diffusion model $g$ and a DAT component. We present the details of each of them below, with more details in Supplementary.

\textbf{CNN Backbone} $\boldsymbol{\phi_I}$\textbf{.} 
We follow previous works \cite{cho2022cross,lin2021mesh,lin2021end} to adopt HRNet \cite{wang2020deep} as our CNN backbone $\phi_I$, that extracts a context feature $f_c \in \mathbb{R}^{2048 \times 7 \times 7}$ from the input image $I$, which is sent into the pose estimator head $\phi_P$.
Moreover, to produce an image feature $f_I \in \mathbb{R}^{128 \times 49}$ to feed into diffusion model $g$, we also flatten $f_c$ and send it into an average pooling layer.

\textbf{Pose Estimator Head} $\boldsymbol{\phi_P}$
is used to initialize the prior distribution $U$, and is a lightweight module consisting of three de-convolutional layers.
$\phi_P$ takes in context features $f_c  \in \mathbb{R}^{2048 \times 7 \times 7}$ extracted from the CNN backbone $\phi_I$, and generates an $xy$ heatmap $E_{x,y} \in \mathbb{R}^{J \times 56 \times 56}$ and a depth heatmap $E_z \in \mathbb{R}^{J \times 56 \times 56}$, where $J$ is the number of joints. 
Since a heatmap is naturally a distribution, we initialize a 3D pose distribution $U$ based on these heatmaps to guide the reverse diffusion process.

\textbf{Diffusion Model $\boldsymbol{g}$} 
facilitates the step-by-step denoising during the reverse process, as shown in the red block of Fig.~\ref{fig:overall_pipeline}.
At the start of step $k$, we are given a noisy 3D mesh input $h_k \in \mathbb{R}^{V \times 3}$ where $V$ is the number of vertices.
We first encode each $v$-th vertex ID to an embedding $E^v_{ID} \in \mathbb{R}^{64}$ via a linear layer and generate a diffusion step embedding $E^k_d \in \mathbb{R}^{61}$ for the $k$-th step via the sinusoidal function. 
Then, we construct $V$ tokens, where each token $x_v \in \mathbb{R}^{128}$ represents the $v^{th}$ vertex, and each token is constructed by concatenating $E^v_{ID}$, $E^k_d$, and the 3D coordinates of the $v$-th vertex in $h_k$. 
These tokens are sent into our diffusion network $g$, which consists of a single vertex self-attention layer, a single vertex-image cross-attention layer, and a linear layer. Refer to Supplementary for more details.

\textbf{Distribution Alignment Technique.}
In this paper, we define the mesh-to-pose function $f$ in DAT as a linear operation (i.e., a matrix) defined using the SMPL model \cite{loper2015smpl} when handling human body mesh (or MANO model \cite{romero2017mano} when handling hand mesh), that regresses 3D joint locations from the estimated mesh sample ($\hat{h}_0$).

\subsection{Training}
\label{sec:training}

\textbf{Learning Reverse Diffusion Process.}
As introduced in previous sections, our diffusion model $g$ is optimized to iteratively denoise $H_{K}$ to get $H_{0}$, i.e., $H_{K} \rightarrow H_{K-1} \rightarrow... \rightarrow H_{0}$.
To achieve this, we first generate ``ground truth" intermediate distributions  $\{H_{1}, H_{2}, ..., H_{K-1} \}$ via the \textit{forward diffusion process}, where we take a ground truth mesh distribution $H_0$ and gradually add noise to it based on Eq.~\ref{eq: original diffusion linear function}.
Then, during model training, we follow previous works \cite{ho2020denoising,song2021denoising} to formulate \textit{diffusion reconstruction loss} $\mathcal{L}_{Diff}$ as follows:
$\mathcal{L}_{Diff}  = \sum_{k=1}^{K} 
\norm{({h}_{k-1} - {h}_{k}) - g ({h}_{k}, k, f_I)}_2^2$,
where $h_k \sim H_k$ and $h_{k-1} \sim H_{k-1}$.

\textbf{Learning Mesh Geometry.} 
To recover an accurate human mesh, we also optimize our diffusion model with the geometric constraints of human mesh. 
Specifically, at each diffusion step, we obtain the estimate of diffusion target $\hat{h}_0$ via Eq.~\ref{eq: estimate hat h_0} and then follow previous work \cite{choi2020pose2mesh} to optimize our model via 4 kinds of losses to constrain the mesh geometry: 
3D Vertex Regression Loss $\mathcal{L}_v$, 3D Joint Regression Loss $\mathcal{L}_j$, Surface Normal Loss $\mathcal{L}_{n}$, and Surface Edge Loss $\mathcal{L}_{e}$. 
See Supplementary for more details.

\textbf{Total Loss.} Combining the losses described above, we define the total loss for training our HMDiff framework as follows:
$\mathcal{L}_{total} = \mathcal{L}_{Diff} + \lambda _{v} \mathcal{L}_{v} + \lambda _{j} \mathcal{L}_{j} + \lambda _{n} 
 \mathcal{L}_{n} + \lambda _{e}  \mathcal{L}_{e}$.

\section{Experiments}

\textbf{Datasets.}
We follow previous works \cite{cho2022cross,lin2021mesh,lin2021end} to evaluate our method on the following datasets.
\textbf{3DPW} \cite{von2018recovering} consists of outdoor images with both 2D and 3D annotations. The training set has 22K images, and the test set has 35K images.
\textbf{Human3.6M} \cite{ionescu2013human3} is a large-scale indoor dataset that has 3.6M images labelled with 2D and 3D annotations. 
Following the setting in previous works \cite{lin2021end,kolotouros2019convolutional,kanazawa2018end},  we train our models using subjects S1, S5, S6, S7 and S8 and test using subjects S9 and S11.
\textbf{FreiHAND} \cite{zimmermann2019freihand} consists of hand actions with 3D annotations.
It has approximately 32.5K training images and 3.9K testing images.

Specifically, we follow previous works \cite{cho2022cross,lin2021mesh,lin2021end} to first train our model with the training sets of Human3.6M \cite{ionescu2013human3}, UP-3D \cite{lassner2017unite}, MuCo-3DHP \cite{mehta2018single}, COCO \cite{lin2014microsoft} and MPII \cite{andriluka20142d}, and then evaluate the model on Human3.6M. Moreover, we follow \cite{cho2022cross,lin2021mesh,lin2021end} to fine tune the model on 3DPW \cite{von2018recovering} training set and evaluate our model on its test dataset. For FreiHAND \cite{zimmermann2019freihand}, following \cite{lin2021end}, we optimize our model on its training set and test our model on its evaluation set.

\textbf{Evaluation Metrics.}
We follow the evaluation metrics from previous works \cite{lin2021end,kolotouros2019convolutional,kanazawa2018end,lin2021mesh,choi2020pose2mesh,cho2022cross}.
Mean-Per-Vertex-Error (\textbf{MPVE}) \cite{pavlakos2018learning} measures the Euclidean distance (in mm) between the predicted vertices and the ground truth vertices.
Next, Mean-Per-Joint-Position-Error (\textbf{MPJPE}) \cite{ionescu2013human3} is a
metric for evaluating human 3D pose \cite{kolotouros2019learning,kanazawa2018end,choi2020pose2mesh}, and measures the Euclidean distance (in mm) between the predicted joints and the ground truth joints.
\textbf{PA-MPJPE}, or Reconstruction Error  \cite{zhou2018monocap}, measures MPJPE after using Procrustes Analysis (PA) \cite{gower1975generalized} to perform 3D alignment.
We also report \textbf{PA-MPVPE} on FreiHAND, which measures MPVE after performing 3D alignment with PA.
On FreiHAND, we also report the F-score \cite{knapitsch2017tanks}, which is the harmonic mean of recall and precision between two sets of points, given a specified distance threshold. Following previous works \cite{lin2021end,lin2021mesh,choi2020pose2mesh,cho2022cross}, we report the F-score at 5mm and 15mm (\textbf{F@5 mm} and \textbf{F@ 15mm}) to evaluate accuracy at fine and coarse scales respectively.

\textbf{Implementation Details.} 
During training, we set the total number of diffusion steps ($K$) at 200 and generate the decreasing sequence $\alpha_{1:200}$ via a linear interpolation function (more details in Supplementary).
We also set the number of samples $N$ to 25.
Following previous works \cite{cho2022cross,lin2021mesh,lin2021end}, we obtain the coarse human mesh with 431 vertices from the original SMPL human mesh (or a human hand mesh with 195 vertices from the original MANO hand mesh) via a GCN model \cite{ranjan2018generating} for training.
We set the learning rate at $0.0001$ and adopt the Adam optimizer to optimize our diffusion model $g$.
{The models $\phi_I$ and $\phi_P$ are pre-trained and then frozen during training of $g$.}
For DAT, we set the activation threshold $r$ at 0.05, and $\gamma$ at 0.2.
During testing, we adopt the DDIM acceleration technique \cite{song2021denoising}, and take 40 steps to complete the whole reverse diffusion process instead of 200.

\subsection{Comparison with State-of-the-art Methods}
We compare our method with existing state-of-the-art HMR methods on Human3.6M and 3DPW datasets.
As shown in Tab.~\ref{tab: main mesh results}, our proposed method can outperform previous works on all metrics. 
This shows that our proposed method (i.e., HMDiff framework with DAT) can effectively recover a high-quality human mesh from a single image.

To further demonstrate the capability of our model to tackle hand mesh reconstruction, we also conduct experiments on FreiHAND dataset \cite{zimmermann2019freihand}. 
As shown in Tab.~\ref{tab: main hand results}, our method outperforms previous state-of-the-art methods, showing its generalizability in this setting as well.

\begin{table}[ht]
\scriptsize
\centering
\vspace{-0.5mm}
\caption{Comparison results on 3DPW and Human3.6M.}
\tabcolsep=1mm
\resizebox{\linewidth}{!}{%
\begin{tabular}{@{}l|ccc|cc@{}}    
\hline
& \multicolumn{3}{c|}{3DPW}  &  \multicolumn{2}{c}{Human3.6M} \\
Method &   MPVE$\downarrow$ & MPJPE$\downarrow$ & PA-MPJPE$\downarrow$ & MPJPE$\downarrow$ & PA-MPJPE$\downarrow$ \\
\hline
Kanazawa et al. \cite{kanazawa2018end} & - & - & 81.3 & 88.0 & 56.8 \\
GraphCMR \cite{kolotouros2019convolutional} & - & - & 70.2 & - & 50.1 \\
SPIN \cite{kolotouros2019learning} & 116.4 & - & 59.2 & - & 41.1 \\
Pose2Mesh \cite{choi2020pose2mesh} & - & 89.2 & 58.9 & 64.9 & 47.0 \\
I2LMeshNet \cite{moon2020i2l} & - & 93.2 & 57.7 & 55.7 & 41.1 \\
VIBE \cite{kocabas2020vibe} & 99.1 & 82.0 & 51.9 & 65.6 & 41.4 \\
METRO \cite{lin2021end} &  88.2 & 77.1 & 47.9 & 54.0 & 36.7 \\
Mesh Graphormer \cite{lin2021mesh} & 87.7 & 74.7 & 45.6 & 51.2 & 34.5 \\
FastMETRO \cite{cho2022cross} & 84.1 & 73.5 & 44.6 & 52.2 & 33.7  \\
\hline
Ours & \textbf{82.4} & \textbf{72.7} & \textbf{44.5} & \textbf{49.3} & \textbf{32.4} \\
\hline
\end{tabular}
}
\label{tab: main mesh results}
\vspace{-5.5mm}
\end{table}

\begin{table}[ht]
\scriptsize
\centering
\caption{Comparison results on FreiHAND. The results with an asterisk (*) are reported by \cite{choi2020pose2mesh}.}
\tabcolsep=1mm
\resizebox{\linewidth}{!}{%
\begin{tabular}{@{}l|cccc@{}}        
\hline
Method &   PA-MPVPE$\downarrow$ & PA-MPJPE$\downarrow$ & F@5 mm$\uparrow$ & F@15 mm$\uparrow$ \\
\hline
Hasson et al. \cite{hasson2019learning} * & 13.2 & - & 0.436 & 0.908 \\
Boukhayma et al. \cite{boukhayma20193d} * & 13.0 & - & 0.435 & 0.898 \\
FreiHAND \cite{zimmermann2019freihand} * & 10.7 & - & 0.529  & 0.935\\
Pose2Mesh \cite{choi2020pose2mesh} & 7.8 & 7.7 & 0.674 & 0.969 \\
I2LMeshNet \cite{moon2020i2l} & 7.6 & 7.4 & 0.681 & 0.973 \\
METRO \cite{lin2021end} &  6.7 & 6.8 & 0.717 & 0.981 \\
Mesh Graphormer \cite{lin2021mesh} & 5.9 & 6.0 & 0.764 & 0.986 \\
FastMETRO \cite{cho2022cross} & - & 6.5 & - & 0.982  \\
\hline
Ours  & \textbf{5.7} & \textbf{5.6} & \textbf{0.781} & \textbf{0.986} \\
\hline        
\end{tabular}
}
\label{tab: main hand results}
\vspace{-2.5mm}
\end{table}

\subsection{Ablation Study}

We also conduct extensive ablation experiments on 3DPW. 
\textbf{See Supplementary for more experiments and analysis.}

\textbf{Impact of Diffusion Process.}
We evaluate the efficacy of the diffusion process by comparing against two baseline models: 
(1) \textbf{Baseline A} possesses the same structure as our diffusion network, but the mesh recovery is conducted in a single step without diffusion.
(2) \textbf{Baseline B} has nearly the same architecture as our diffusion network, but we stack the network multiple times to approximate the computational cost of our method. 
We remark that both baselines are optimized to directly predict the human mesh with a forward pass, instead of learning the step-by-step reverse diffusion process.
We report the results of our proposed method and the baselines in Tab.~\ref{table: Ablation Diffusion}.
The performance of both baselines are much worse than ours, suggesting that much of the performance improvement comes from the designed diffusion pipeline.

\setlength{\intextsep}{0pt}%
\setlength{\columnsep}{6pt}%
\begin{table}[h]
\centering
\scriptsize
\caption{Evaluation of diffusion pipeline.}
\label{table: Ablation Diffusion}
\begin{tabular}{l|ccc}
\hline
Method & MPVE & MPJPE & PA-MPJPE\\
\hline
Baseline A & 104.0 & 93.2 & 57.6 \\
Baseline B & 97.1 & 85.9 & 52.7  \\
\hline
Ours & 82.4  & 72.7 & 44.5  \\
\hline
\end{tabular}
\vspace{-4.5mm}
\end{table}

\textbf{Impact of DAT.}
We also verify the impact of our DAT by comparing against three baselines: 
(1) \textbf{Standard} where the standard diffusion process is used (without DAT), with image feature $f_I$ as an input feature.
(2) \textbf{Pose Feature} where the standard diffusion process is used (without DAT), with pose heatmap $U$ as an input feature.
(3) \textbf{Both Features} where we concatenate the pose heatmap with the image features $f_I$ before feeding them to the diffusion model $g$.
For each method, we tune the DDIM \cite{song2021denoising} acceleration rate to perform the reverse diffusion process with different diffusion steps, and report the number of steps used where they obtain the best performance. 
Results are reported in Tab.~\ref{table: Ablation DRT}, where our method significantly outperforms all baselines while using fewer steps.
This is because, as compared to these baselines, our DAT can explicitly constrain the initial steps to align towards the prior distribution, which improves the performance as well as the speed of convergence to the target $H_0$.

\begin{table}[h]
\vspace{1.5mm}
\centering
\scriptsize
\caption{Evaluation of impact of DAT.}
\label{table: Ablation DRT}
\begin{tabular}{l|cccc}
\hline
Method & MPVE & MPJPE & PA-MPJPE & Steps needed \\
\hline
Standard & 94.9 & 83.3 & 51.3 & 200  \\
Pose Feature & 92.7 & 81.4 & 50.5 &  200 \\
Both Features & 92.3  & 81.1 & 49.9 & 100 \\
\hline
Ours (with DAT) & 82.4  & 72.7 & 44.5 & 40 \\
\hline
\end{tabular}
\vspace{2.5mm}
\end{table}

\textbf{Impact of DAT Components.}
We study the impact of various DAT components by comparing against the following:
(1) \textbf{Ours w/} $\boldsymbol{H_U}$ starts the reverse process from a noisy distribution $H_U$, generated by upsampling the prior pose distribution $U$ to a mesh distribution.
(2) {\textbf{Disrupted} baseline} directly aligns $h_k$ to the prior pose distribution $U$ via an $L_2$ loss which directly modifies $h_k$ (i.e., disrupts the diffusion process).
(3) \textbf{Ours w/ \boldm{${u_c}$}} uses a single 3D pose $u_c$ as prior knowledge (detected from the heatmap), instead of the 3D pose distribution $U$.
(4) \textbf{Ours w/o AS} activates DAT over the whole reverse process, i.e., without using the Activation Strategy.
As shown in Tab.~\ref{table: Ablation DRT component}, our method outperforms all baselines, showing the efficacy of our design choices.

\begin{table}[h]
\centering
\scriptsize
\caption{Evaluation of DAT components.}
\label{table: Ablation DRT component}
\begin{tabular}{l|ccc}
\hline
Method & MPVE & MPJPE & PA-MPJPE\\
\hline
Ours w/ $H_U$ & 93.5  & 82.1 & 50.4\\
Disrupted & 91.3 & 78.0 & 50.0\\
Ours w/ $u_c$ & 89.9  & 76.5 & 49.2 \\
Ours w/o AS  & 85.3 & 72.9 & 46.6 \\
\hline
Ours & 82.4  & 72.7 & 44.5  \\
\hline
\end{tabular}
\end{table}

\begin{figure*}[t]
  \centering
  \includegraphics[width=0.9\linewidth]{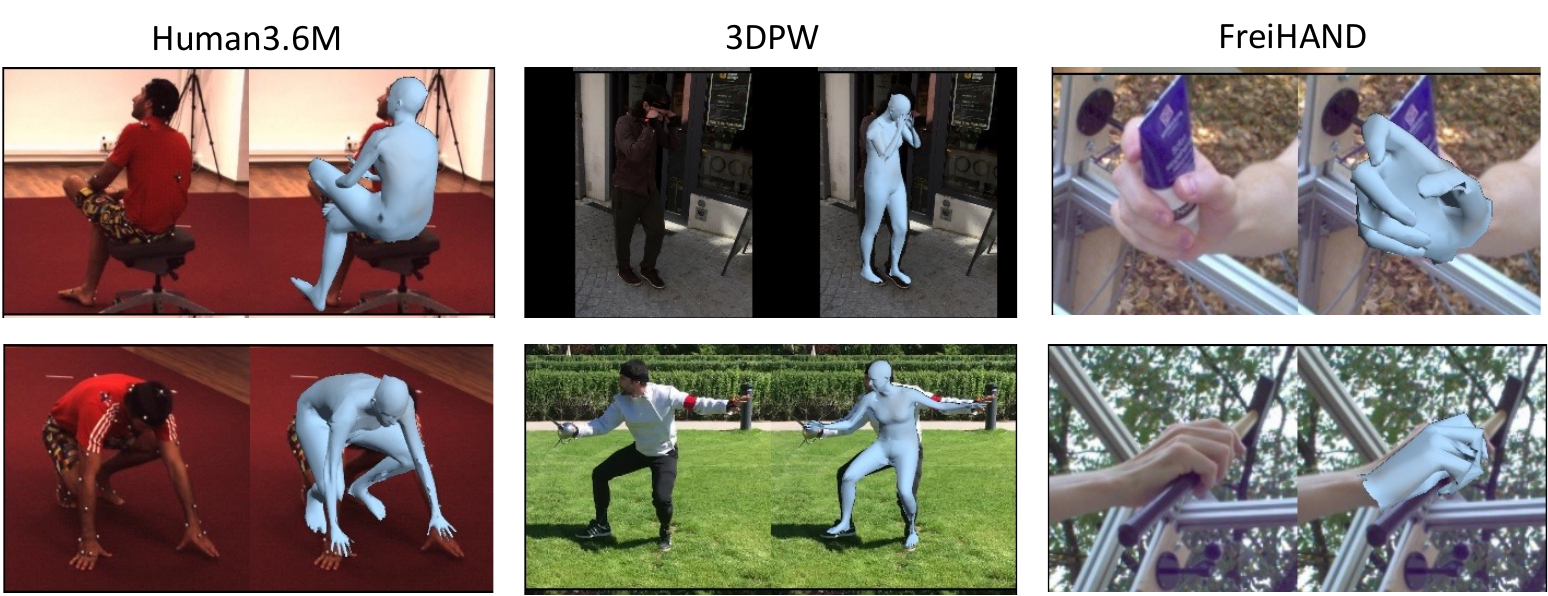}
  \caption{
  Visualization of body and hand mesh outputs using our method. Our method effectively recovers the mesh even under ambiguity (e.g., due to heavy occlusions), and produces high-quality results.
  (More visualization results are shown in Supplementary.)
    }
  \label{fig:visualization_mesh}
\end{figure*}

\begin{figure}[h]
  \centering
  \includegraphics[width=\linewidth]{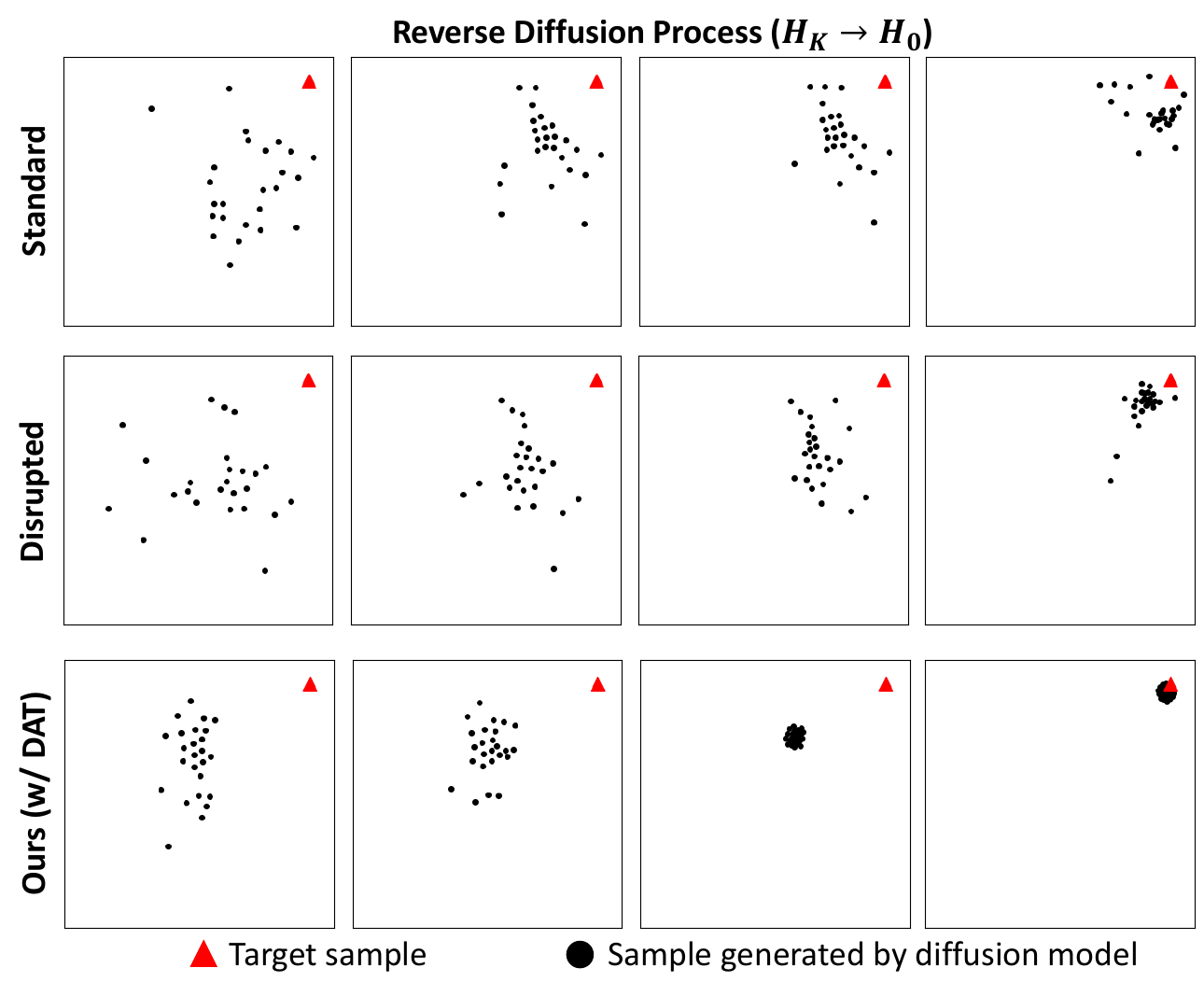}
\caption{  Visualization of mesh distributions generated by different variations of the reverse diffusion process.   
  We observe that the mesh distribution produced by our method (bottom) smoothly converges towards the target. 
  }  
\label{fig:visualization_DAT}
\vspace{-2.5mm}
\end{figure}

\textbf{Visualization of the Mesh Distribution Diffusion Process.}
We visualize three variations of the diffusion process: 
(1) \textbf{Standard} baseline where we perform standard diffusion without any alignment with prior distribution $U$.
(2) \textbf{Disrupted} baseline where we directly modify the samples $h_k$ to be closer to prior distribution $U$ at every step via an $L_2$ loss. 
(3) \textbf{Ours (w/ DAT)}, which is our method.
To visualize the mesh distributions, we map the $N$ mesh samples to 2D space via the t-SNE algorithm \cite{van2008visualizing}, where each dot represents a sample.   
As shown in the top row of Fig.~\ref{fig:visualization_DAT}, the generated samples of the Standard baseline are spread widely apart, showing that the mesh distribution does not converge to the target well.
The Disrupted baseline receives prior distribution information during the diffusion process, and thus converges better than the Standard baseline, but the diffusion process is disrupted and thus the final mesh distribution deviates from the target (as shown in the middle row of Fig.~\ref{fig:visualization_DAT}).
In contrast, the mesh distribution produced by our method smoothly converges towards the target (as shown in the bottom row of Fig.~\ref{fig:visualization_DAT}).
For quantitative results, refer to Tab.~\ref{table: Ablation DRT} and Tab.~\ref{table: Ablation DRT component}.

\textbf{Visualization of Samples.}
Some qualitative results of our method on Human3.6M and 3DPW are shown in Fig.~\ref{fig:visualization_mesh}. 
We observe that our method can handle challenging cases, e.g., under heavy occlusions or with noisy background, showing its
strong ability to handle uncertainty.

\textbf{Inference Speed.}
Here, we conduct experiments on a single GeForce RTX 3090 card and compare the speed of our proposed method (HMDiff with DAT) with existing methods in terms of the accuracy metric (MPJPE) and inference speed (FPS) in Tab.~\ref{table: Ablation speed}.  
Our method can achieve a competitive speed compared with the current SOTA \cite{cho2022cross} while significantly outperforming it (as shown in Tab.~\ref{tab: main mesh results} and Tab.~\ref{tab: main hand results}).

\setlength{\intextsep}{0pt}%
\setlength{\columnsep}{6pt}%
\begin{table}[h]
\centering
\scriptsize
\caption{Comparison of inference speed.}
\label{table: Ablation speed}
\begin{tabular}{l|cc}
\hline
Method &  Human3.6M(MPJPE)  & FPS \\
\hline
\cite{cho2022cross} & 52.2 & 23  \\
\hline
Ours & 49.3 & 18  \\
\hline
\end{tabular}
\end{table}

\section{Conclusion}

We present HMDiff, a novel diffusion-based framework that frames mesh recovery as a reverse diffusion process to tackle monocular HMR.
We infuse prior distribution information via our proposed DAT during the diffusion process to obtain improved performance.
Extensive experiments show that the proposed method achieves state-of-the-art performance on three widely used benchmark datasets.

\section{Acknowledgements}
This work was supported by the Singapore Ministry of Education (MOE) AcRF Tier 2 under Grant MOE-T2EP20222-0009, the National Research Foundation Singapore through AI Singapore Programme under Grant AISG-100E-2020-065, and SUTD SKI Project under Grant SKI 2021\_02\_06.

\end{document}